\documentclass[11pt]{article}
\usepackage[margin=1in]{geometry}
\usepackage{amsmath,amssymb,amsthm,mathtools,booktabs,graphicx,array,tabularx,microtype,natbib,enumitem}
\usepackage[hidelinks]{hyperref}
\usepackage{orcidlink}
\usepackage[T1]{fontenc}
\usepackage{authblk}
\usepackage{tikz}
\usetikzlibrary{positioning,arrows.meta}
\newtheorem{theorem}{Theorem}
\newcommand{\PMD}{\operatorname{PMD}}\newcommand{\softmax}{\operatorname{softmax}}\newcommand{\KL}{D_{\mathrm{KL}}}\newcommand{\F}{\mathcal F}\newcommand{\E}{\mathbb E}\newcommand{\one}{\mathbf 1}\newcommand{\norm}[1]{\lVert#1\rVert}\newcommand{\ip}[2]{\langle#1,#2\rangle}
\title{PolicyAttention: Softmax Attention Implements Policy Mirror Descent for Closed-Loop Control}
\author[1,2]{Yuhe Sui\,\orcidlink{0009-0002-7456-0804}}
\author[1,3]{Yingzhi Tang\,\orcidlink{0009-0004-7571-0100}}
\author[1,4]{Shufang Chen\,\orcidlink{0009-0009-3627-7726}}
\affil[1]{Quantitative Research Society, Singapore}
\affil[2]{Nanyang Technological University, Singapore}
\affil[3]{The Chinese University of Hong Kong, Shenzhen, China}
\affil[4]{The University of Hong Kong, Hong Kong SAR, China}

\begin{document}
\maketitle
\begin{abstract}
A Transformer can match a reinforcement-learning update on fixed inputs yet fail once its own returned policies alter subsequent inputs. We study this closed-loop gap for negative-entropy policy mirror descent (PMD), whose update is softmax-native: $\PMD_\eta(\pi,Q)=\softmax(\log\pi+\eta Q)$. The reference recursion is known Q-TD-PMD; our contribution is its causal implementation and learned closed-loop interface. A fixed causal-softmax decoder realizes an inexact actor--environment--one-step-critic loop with explicit actor, routing, and sampling residuals, and a returned-policy theorem propagates those errors to the policy actually output, with no unused final critic residual. A finite-cap pre-LayerNorm/final-LayerNorm compilation and a scoped effective-coordinate result connect the construction to standard Transformer operations and the PMD-aligned training objective.

Empirically, a separately trained pre-LN Transformer is closest to PMD among fixed update hypotheses, satisfies a preregistered repeated-control criterion on five fresh runs, and retains that behavior under no-retraining environment shifts. At larger retrained sizes, PolicyAttention attains median normalized returned-policy losses $0.0193$ at $S=8$ and $0.0273$ at $S=16$. Under the common scoring harness, these are $18$--$28\times$ lower than our qualified Liang--Lai and Algorithm Distillation adaptations. PolicyAttention uses an exact one-step Bellman backup with model access, whereas the external adaptations update from sampled interactions; this is therefore a common-harness performance separation rather than an information-matched comparison. Together, the results distinguish local algorithmic fidelity, closed-loop reliability, and task performance, and show how softmax attention can serve as the policy-improvement operator itself.
\end{abstract}

\section{Introduction}
Transformers can execute learning algorithms in context, but an adaptive controller faces a stronger test than a static predictor: its current output changes the distribution of its future inputs. In reinforcement learning (RL), a returned policy changes successor actions, critic targets, and the context processed on the next round. A model can therefore look algorithmic for one step and cease to be useful when repeatedly deployed.

PolicyAttention asks whether \emph{standard causal softmax attention can implement and learn a genuine policy-improvement algorithm that remains effective after its own policies recursively alter the control inputs}. Negative-entropy PMD is a natural target because its statewise update is exactly
\begin{equation}
\PMD_\eta(\pi,Q)(a\mid s)=\frac{\pi(a\mid s)e^{\eta Q(s,a)}}{\sum_b\pi(b\mid s)e^{\eta Q(s,b)}}=\softmax(\log\pi+\eta Q)_a.
\label{eq:pmd}
\end{equation}
The identity and the associated PMD/TD-PMD control recursions are prior RL methodology \citep{xiao2022,johnson2023,geist2019,lan2023,liuliwei2025}. Our object is the implementation layer: a causal softmax realization with measurable residuals, a returned-policy interface for those residuals, and a learned closed-loop test.

\begin{figure}[t]
\centering
\resizebox{.98\textwidth}{!}{%
\begin{tikzpicture}[
  >=Latex,
  font=\small,
  box/.style={draw=black!45, rounded corners=2pt, thick, align=center, inner sep=5pt, minimum height=8mm, fill=white},
  theory/.style={box, fill=red!4, draw=red!55},
  stage/.style={box, fill=black!1},
  critic/.style={box, fill=green!4, draw=green!45}
]
\node[stage, minimum width=2.1cm] (ctx) at (1.4,0.7) {\shortstack{policy + value\\$(\pi_k,Q_k)$}};
\node[theory, minimum width=2.95cm, right=0.65cm of ctx] (actor) {\shortstack{softmax PMD actor\\$\softmax(\log \pi + \eta Q)$}};
\node[stage, minimum width=2.0cm, right=0.65cm of actor] (env) {\shortstack{environment\\interaction}};
\node[critic, minimum width=2.45cm, right=0.65cm of env] (critic) {\shortstack{one-step critic\\$\mathcal{F}^{\pi_{k+1}}Q_k$}};
\node[stage, minimum width=2.15cm, right=0.65cm of critic] (next) {\shortstack{next context\\$(\pi_{k+1},Q_{k+1})$}};
\draw[->, thick, black!55] (ctx) -- (actor);
\draw[->, thick, black!55] (actor) -- (env);
\draw[->, thick, black!55] (env) -- (critic);
\draw[->, thick, black!55] (critic) -- (next);
\end{tikzpicture}%
}

\vspace{0.5em}
\setlength{\tabcolsep}{5pt}
\renewcommand{\arraystretch}{1.08}
\small
\begin{tabularx}{.98\textwidth}{>{\raggedright\arraybackslash}p{.12\textwidth}>{\raggedright\arraybackslash}X>{\raggedright\arraybackslash}X>{\raggedright\arraybackslash}X}
\toprule
& \textbf{Theory} & \textbf{Learned controller} & \textbf{Literature-facing comparison}\\
\midrule
Focus & fixed causal-softmax PMD/critic implementation; explicit actor, routing, sampling, and returned-policy residuals & four-layer pre-LN Transformer; PMD-aligned objective; exact-critic repeated-control test plus descriptive learned-critic evidence & PolicyAttention vs Liang--Lai and Algorithm Distillation; Exact PMD is the oracle; learned PMD-target model is calibration only\\
\bottomrule
\end{tabularx}
\normalsize
\caption{\textbf{One mechanism-to-control chain.} The causal loop is shown above; the table below separates the constructive theorem, the learned-controller evidence, and the literature-facing benchmark. Exact PMD is the non-learned oracle, while the learned PMD-target model serves only as calibration.}
\label{fig:overview}
\end{figure}
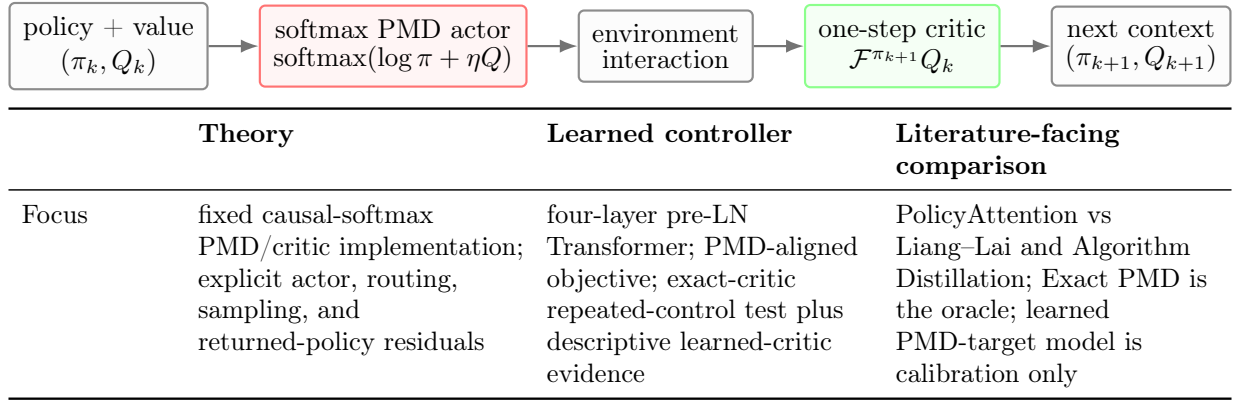

The contribution is one chain. \textbf{First}, a fixed two-block/two-head causal decoder realizes PMD followed by one-step policy evaluation under the policy it actually returned, with explicit finite routing and sampling errors. \textbf{Second}, these errors feed a last-policy theorem that controls the final returned policy and correctly omits the unused final critic residual. \textbf{Third}, a specially compiled pre-LN/final-LN construction and reverse-KL/Fisher effective-coordinate result explain how the same target computation fits standard normalized operations and the PMD-aligned objective. \textbf{Fourth}, fresh experiments show the learned PMD-like actor survives adaptive repetition and fixed-shape distribution shifts. \textbf{Finally}, a qualified common-harness benchmark compares PolicyAttention with a matched PMD-target reference Transformer and two published-method adaptations, placing the mechanism result in a literature-facing performance context.

\section{Related Work}
Transformers have been shown to implement or learn in-context regression and gradient procedures \citep{garg2022,akyurek2023,vonoswald2023}. In RL, Wang et al. learn TD-style policy evaluation in context \citep{wang2025}; Xie et al. give a standard-softmax construction for weighted softmax TD and policy evaluation \citep{xie2026}. Liang and Lai construct semi-gradient SARSA and actor--critic updates with linear self-attention and analyze teacher-mimicking learning \citep{lianglai2026}. Algorithm Distillation (AD) trains a causal sequence model on complete learning histories and obtains empirical in-context improvement without weight updates \citep{laskin2023}; S5RL uses structured state-space models in a different online-PPO/meta-RL paradigm \citep{lu2023}.

\begin{table}[t]\centering\scriptsize
\caption{Source-verified capability map. ``Softmax'' denotes standard softmax attention in the relevant construction/model; PolicyAttention's final column denotes the returned-policy residual interface introduced here.}
\begin{tabular}{@{}lccccc@{}}\toprule
Work & Softmax & Eval. & Control & Learned & Return-error \\\midrule
Wang et al. 2025 & emp. & \checkmark & -- & \checkmark & --\\
Xie et al. 2026 & \checkmark & \checkmark & -- & \checkmark & --\\
Liang--Lai 2026 & linear & \checkmark & \checkmark & \checkmark & --\\
Algorithm Distill. & \checkmark & implicit & \checkmark & \checkmark & --\\
PolicyAttention & \checkmark & \checkmark & \checkmark & \checkmark & \checkmark\\\bottomrule
\end{tabular}
\label{tab:cap}
\end{table}

The gap is not ``control versus no control.'' It is the intersection of \emph{standard softmax policy improvement, explicit causal implementation errors, and returned-policy closed-loop accounting}. The benchmark numbers for Liang--Lai and AD below are ours from faithful common-harness adaptations, not results reported by those papers.

\section{Causal softmax realization}
Let $\mathcal M=(\mathcal S,\mathcal A,P,r,\gamma)$ be finite with $|r|\le R_{\max}$ and $B=R_{\max}/(1-\gamma)$. Define the one-step policy-evaluation and optimality operators
\begin{align}
(\F^\pi Q)(s,a)&=r(s,a)+\gamma\E_{S'\mid s,a}\ip{\pi(\cdot\mid S')}{Q(S',\cdot)},\\
(\F Q)(s,a)&=r(s,a)+\gamma\E_{S'\mid s,a}\max_bQ(S',b).
\end{align}
At round $k$, the implemented actor returns $\widehat\pi_{k+1}$ from $(\widehat\pi_k,Q_k)$, then the environment generates critic data under that returned policy. Define same-context residuals
\begin{align}
\zeta_k&=\max_s\norm{\widehat\pi_{k+1}(\cdot\mid s)-\PMD_{\eta_k}(\widehat\pi_k,Q_k)(\cdot\mid s)}_1,\\
\delta_k&=\norm{Q_{k+1}-\F^{\widehat\pi_{k+1}}Q_k}_\infty.
\label{eq:resids}
\end{align}
These are local implementation quantities, not task losses after two trajectories diverge.

\paragraph{Construction.} The normalization-free circuit has two blocks, two heads, head dimension $S+A+3$, payload width $6S+6A+14$, a causal mask, no positional embeddings, and a role-gated ReLU map. Actor memories carry state/action identity, $\log\widehat\pi_k$, and $Q_k$; intended attention scores are a common state offset plus $\log\widehat\pi_k(a\mid s)+\eta_kQ_k(s,a)$, so intended-group softmax is exactly PMD. Finite off-group mass yields
\begin{equation}
\zeta_k\le\frac{2R_k^{\rm act}}{1+R_k^{\rm act}},\quad
R_k^{\rm act}\le(S-1)e^{-\kappa+2\eta_kB}+Se^{-(\kappa+\nu)+\eta_kB}.
\end{equation}
The critic uses \emph{lookup $\to$ local target $\to$ predecessor average}. With balanced fresh packets, lookup/aggregation leakage masses $\lambda_1,\lambda_2$ and $m$ samples per predecessor give
\begin{equation}
\delta_k\le2B(\gamma\lambda_1+\lambda_2)+B\sqrt{\frac{2\log(2SAK/\alpha)}{m}}
\label{eq:critic}
\end{equation}
simultaneously over the declared finite calls. Resource certificates precede weight choice; the frozen decoder is then reused across admissible MDPs and histories.

\begin{theorem}[Finite-horizon causal realization]
Fix finite $S,A$, reward and horizon caps, deterministic step/packet caps, routing tolerances, and confidence $1-\alpha$. One fixed causal-softmax decoder can be selected before the MDP and realized history so that repeated actor/environment/critic calls realize inexact Q-TD-PMD with actor residual bounded as above and critic residual satisfying~\eqref{eq:critic} on the declared common event.
\end{theorem}

\section{Returned-policy control and normalized compilation}
Let $g_k(s)=\max_aQ_k(s,a)-\ip{\widehat\pi_{k+1}(\cdot\mid s)}{Q_k(s,\cdot)}$ and $\bar g_k$ be its transition-lifted maximum. Bellman contraction gives $E_{k+1}\le\gamma E_k+\delta_k+\gamma\bar g_k$ for $E_k=\norm{Q^\star-Q_k}_\infty$.

\begin{theorem}[Returned-policy error interface]\label{thm:last}
For every $T\ge1$,
\begin{align}
\norm{Q^\star-Q^{\widehat\pi_T}}_\infty\le{}&\frac{2\gamma^T E_0}{1-\gamma}+\frac{2}{1-\gamma}\sum_{k=0}^{T-2}\gamma^{T-1-k}\delta_k\\
&+\frac{2}{1-\gamma}\sum_{k=0}^{T-2}\gamma^{T-k}\bar g_k+\frac{\gamma}{1-\gamma}\bar g_{T-1}.
\label{eq:last}
\end{align}
No $\delta_{T-1}$ appears because $\widehat\pi_T$ is formed from $Q_{T-1}$ before an unused final critic backup.
\end{theorem}

PMD optimality also gives $\bar g_k\le\bar D_k/\eta_k+B\zeta_k$, where $\bar D_k$ is the greedy-set prior-mass price. This identifies why low one-step actor error does not by itself certify repeated control.

A specially engineered normalized construction adds a two-coordinate affine carrier. For $\operatorname{Embed}(x)=Cc+Jx$ and an appropriate LayerNorm gain,
$J^\top\operatorname{LN}(Cc+Jx)=\rho_H(x)x$ with $\rho_H(x)=(1+\norm{x}^2/H^2)^{-1/2}$, so sufficiently large finite $H$ makes normalization a controlled perturbation. The actor readout explicitly mixes with uniform, $\widehat p=(1-\varphi)p+\varphi\one/A$, giving a design-time floor and $\norm{\widehat p-p}_1\le2\varphi(A-1)/A$. With constant $\eta=\log(A/\varphi)/\theta$, the returned policy obeys a geometric transient plus residual neighborhood
\begin{equation}
\norm{Q^\star-Q^{\widehat\pi_T}}_\infty\le \frac{(2E_0+4B)\gamma^T}{1-\gamma}+\frac{2\gamma}{(1-\gamma)^2}(\delta+B\zeta+\theta).
\end{equation}
This is a finite-cap real-arithmetic existence result on an engineered slice, not a raw-parameter training theorem.

\subsection{Why the returned-policy indexing matters}
Theorem~\ref{thm:last} is a last-iterate statement rather than a bound on an auxiliary critic. Unrolling the critic recursion only through $Q_{T-1}$ gives
\begin{equation}
E_{T-1}\le \gamma^{T-1}E_0+\sum_{k=0}^{T-2}\gamma^{T-2-k}\delta_k+\sum_{k=0}^{T-2}\gamma^{T-1-k}\bar g_k.
\end{equation}
For $\pi=\widehat\pi_T$, the Bellman resolvent satisfies
$\norm{Q^\star-Q^\pi}_\infty\le 2\gamma E_{T-1}/(1-\gamma)+\gamma\bar g_{T-1}/(1-\gamma)$.
The policy is already determined at this point. Computing $Q_T$ could be useful for another actor call, but it cannot retroactively change $\widehat\pi_T$; charging $\delta_{T-1}$ would therefore measure work not used by the returned policy. This causal accounting is also the reason the empirical certificate audit evaluates realized actor/critic sequences at the horizon actually returned.

\subsection{Support, greediness, and the role of the explicit mixture}
For $G_k(s)=\arg\max_aQ_k(s,a)$ define $D_k(s)=-\log\widehat\pi_k(G_k(s)\mid s)$. Comparing PMD with the current policy restricted to $G_k(s)$ yields
\begin{equation}
\max_aQ_k(s,a)-\ip{\PMD_{\eta_k}(\widehat\pi_k,Q_k)}{Q_k(s,\cdot)}\le D_k(s)/\eta_k.
\end{equation}
Replacing the exact PMD row by the implemented actor costs at most $B\zeta_k$. Thus actor fidelity becomes control-relevant only together with a prior-mass/greedification condition. The explicit uniform mixture in the normalized construction gives the design-time support floor $\widehat p_a\ge\varphi/A$, hence $D_k\le\log(A/\varphi)$ after the first call. This closes the finite-horizon control statement with a constant step, but it changes the executed controller and can impose a performance cost when the unmixed actor would otherwise concentrate more strongly. Section~7 reports that cost empirically rather than treating the floor as automatically beneficial.

\section{What the learned model learns}
For a predicted row $p$, define $\ell_{\rm prox}(p)=\KL(p\|\pi)-\eta\ip{p}{Q}$. If $q=\PMD_\eta(\pi,Q)$, then
\begin{equation}
\ell_{\rm prox}(p)-\ell_{\rm prox}(q)=\KL(p\|q).
\end{equation}
After removing the action-constant softmax gauge, a realizable centered linear effective-logit model has population gradient $\nabla R(\vartheta)=\E[\Phi^\top H(q_\vartheta)\Phi](\vartheta-\vartheta^\star)$, with $H(q)=\operatorname{diag}(q)-qq^\top$. Bounded features and task richness therefore give convergence in these effective coordinates for gradient flow and sufficiently small-step GD. This explains objective selection, not optimization of the full deep Transformer.

\begin{figure*}[t]\centering\includegraphics[width=.82\textwidth]{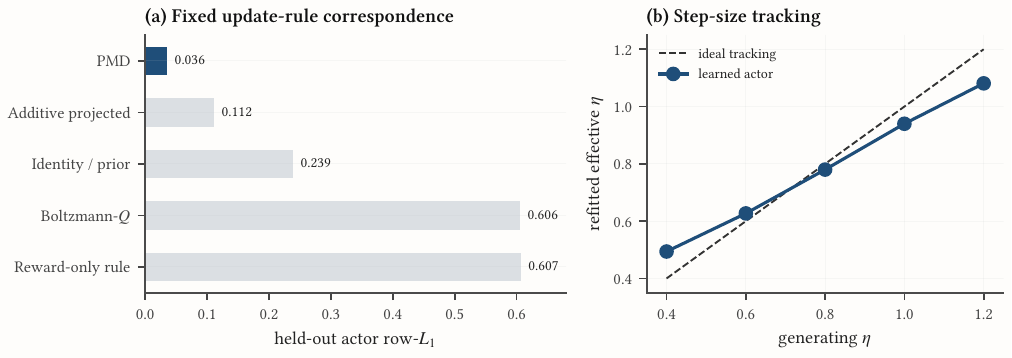}\caption{\textbf{The learned actor exhibits the predicted PMD signature before closed-loop testing.} (a) On prior held-out one-step inputs, PMD has mean row-$L_1$ error $0.0357$, versus $0.1116$ for the nearest fixed alternative. (b) Refitted effective step size changes monotonically with supplied $\eta$, with shrinkage toward the training-range center. These are local mechanism measurements, not returned-policy performance.}\label{fig:mech}\end{figure*}

The evaluated model is a separate four-layer/four-head pre-LN Transformer (width 64, FFN 128, zero dropout, no positional embeddings). Earlier held-out tests report PMD row-$L_1=0.035741$, fitted effective $\eta$ mean $0.783667$, and critic TD MAE $0.121234$. Figure~\ref{fig:mech} shows that PMD is substantially closer than additive-projected, identity, Boltzmann-$Q$, or reward-only fixed rules. The construction and learned network are connected through the target computation and measured residuals, not by parameter identity.

\subsection{Learned architecture and training interface}
The learned system uses a conventional four-layer, four-head pre-LN Transformer with $d_{\rm model}=64$, feed-forward width 128, zero dropout, and no positional embeddings. Actor examples provide structured $(\pi,Q,\eta)$ fields; critic examples provide the fields required for the one-step Bellman target. The PMD-aligned actor is trained through the proximal objective above rather than direct PMD probability labels. In the fresh $S=4$ block, the training range is $\eta\in[0.4,1.2]$ with nominal $0.8$, 24 training MDPs and 2,048 examples per run, AdamW at $3\times10^{-4}$, weight decay $10^{-4}$, batch 64, and 51,200 optimizer steps. Fitted reference steps on the disjoint calibration fixture are $0.800,0.795,0.800,0.795,0.805$, so reference migration is negligible at the primary budget.

The theory and learned experiment deliberately share an \emph{interface}, not parameter values. The constructive decoder has sparse role-specific weights and theorem-scale margins; the learned model has ordinary dense trainable layers. We therefore test the target computation at three levels: same-context actor deviation $\zeta$, critic deviation $\delta$, and returned-policy loss after adaptive composition. This avoids the stronger and unsupported claim that optimization recovers the constructed sparse circuit.

\section{Experimental Design and Estimands}
The empirical program is organized to test implications of the theory rather than to infer the mechanism from reward alone. Table~\ref{tab:ladder} summarizes the ladder. The local actor diagnostic uses the \emph{same} $(\widehat\pi_k,Q_k)$ context for the learned prediction and exact PMD target. The repeated-control experiments then allow the learned and reference trajectories to diverge and score the policies each controller actually returns. Finally, the literature benchmark retains the same exact scoring metric but changes the learned method and, for the external adaptations, the information interface.

\begin{table*}[t]\centering\small
\caption{Claim-to-estimand map. Each row answers a different scientific question; evidence on one row is not substituted for another.}
\begin{tabularx}{.96\textwidth}{@{}p{.16\textwidth}p{.25\textwidth}X@{}}\toprule
Question & Coordinate / comparison & Verified result and interpretation\\\midrule
What actor rule was learned? & same-context row-$L_1$ to fixed update hypotheses & PMD $0.0357$, nearest alternative $0.1116$: strong local PMD correspondence on held-out one-step inputs.\\
Does local correspondence survive feedback? & learned actor + exact one-step critic / Exact PMD oracle & median ratio $1.052$; all five trained runs satisfy the preregistered $1.5$ criterion.\\
Does the fully learned loop remain near the oracle? & learned actor + learned critic / Exact PMD oracle & descriptive median ratio $1.050$ at $S=4$; no reliability margin was registered for this comparison.\\
Does it survive environment shift? & same learned actors + exact critic, four no-retraining families & all four families satisfy the declared criterion separately; the first three use 64 MDPs each; the structured ring is one MDP $\times$ 64 initial policies.\\
Does the phenomenon persist at larger size? & retrained $S=8$ and $S=16$ models & PolicyAttention remains within 1.31$\times$ and 1.64$\times$ the Exact PMD oracle at $T=20$; this is retraining, not zero-shot size generalization.\\
How does it compare with neighboring learned ICRL? & qualified common-harness adaptations & exact-critic PolicyAttention is $17.7$--$28.2\times$ lower-loss at $T=20$; at $S=8$ the learned-critic check remains $20.2$--$24.2\times$ lower but uses $7.2\times$ the adaptations sampling budget.\\
What do matched controls establish? & learned PMD-target and reward-only controls & direct PMD supervision is competitive; reward-only has lower $S=4,T=20$ task loss, so mechanism fidelity and task optimization are distinct.\\\bottomrule
\end{tabularx}\label{tab:ladder}
\end{table*}

\paragraph{Returned-policy loss.}
For each evaluated MDP we compute exact $Q^\pi$ for the controller's returned policy and exact $Q^\star$, then normalize the sup-norm Q-loss by the common initial-policy gap. This produces a dimensionless remaining-loss coordinate shared across methods. Exact PMD is run on the same MDPs as an algorithmic reference trajectory. It is not counted as a trained competitor because its role is to expose the target algorithm's achievable trajectory under model-based exact updates. Every learned method is scored by the same exact evaluator even when its own update rule observes only sampled transitions.

\paragraph{The learned/reference ratio.}
For the fresh confirmatory experiment, each trained run is summarized at $T=20$ by
\begin{equation}
r_j=\frac{\operatorname{median}_{m}L_{j,m}(\mathrm{learned})}{\operatorname{median}_{m}L_m(\mathrm{Exact\ PMD\ Ref.})}.
\end{equation} A ratio below one has lower task loss than the reference on that run; one is parity; one to $1.5$ is worse than the oracle trajectory but remains inside the preregistered usefulness margin. This threshold is an operational reliability criterion fixed before the fresh evidence, not a statistical equivalence margin.

\paragraph{Certificate audit.}
The returned-policy inequality is also recomputed from realized residual sequences at horizons $\{1,2,5,10,20,40\}$. Across the fresh confirmation, breadth, and reward-only evaluations, all 69,120 audited rows satisfy the inequality. The median slack is large (about 10.77), so the audit establishes consistency of indexing and measured residual interfaces; it is not used as a calibrated numerical predictor of task loss.

\section{Closed-Loop Reliability and Scaling}
The fresh confirmation fixed a 51,200-step budget and fresh seeds before execution. The primary comparison replaces only the exact PMD actor with the learned actor while retaining the same exact one-step critic. At $T=20$ the five learned/reference ratios are $0.978,1.396,1.237,0.918,1.052$ (median $1.052$); all five meet the preregistered $\le1.5$ criterion. A hierarchical learned-actor-minus-reference difference has median $0.000177$ with 95\% interval $[-0.000137,0.000748]$: this interval is a different estimand from the predeclared ratio criterion and is not an equivalence test. With the same trained checkpoints and their learned critic, the fully learned loop has descriptive median oracle ratio $1.050$ and full-loop-minus-oracle median $0.000518$ with interval $[0.000008,0.002155]$; no $1.5$ margin was registered for this comparison.

\begin{figure*}[t]\centering\includegraphics[width=.93\textwidth]{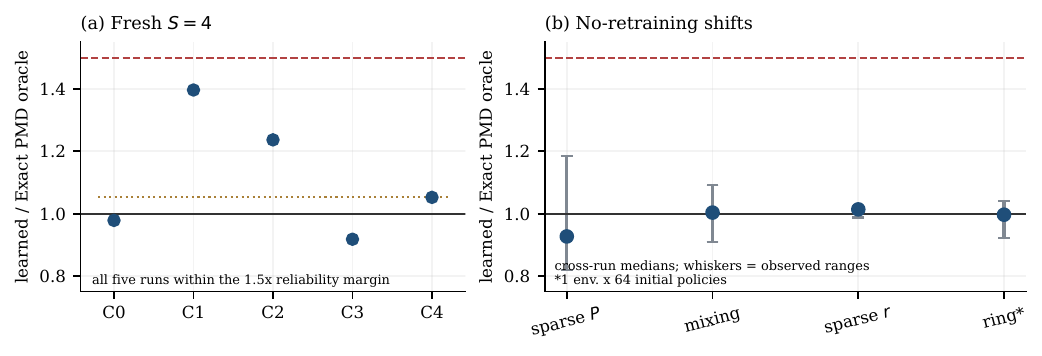}\caption{\textbf{Local PMD correspondence remains useful under repeated deployment.} (a) With an exact one-step critic, all five confirmatory $S=4$ runs stay within the preregistered $1.5\times$ margin to the Exact PMD oracle. (b) The same trained actors satisfy the criterion across four no-retraining shifts; whiskers show observed five-run ranges.}\label{fig:closed}\end{figure*}

Without retraining, all four declared $S=4$ shift families pass separately: cross-run median learned/reference ratios are $0.927$, $1.003$, $1.014$, and $0.997$. The first three families contain 64 distinct environments each; the structured ring is one environment evaluated from 64 initial policies. The breadth result remains positive if the structured-ring family is omitted. Subsequent scale experiments retrain the architecture at larger state counts; these are size-scaling experiments, not zero-shot cardinality generalization.

\subsection{Experimental units and statistical treatment}
The trained model/run is the outer independent unit. The 64 MDPs within a run are paired fixtures used to estimate each trained controller; they are not substitutes for independent training replications. The fresh confirmation uses five trained runs. The larger-size literature benchmark uses three training identities per learned method at each size. Final paired intervals resample runs first and shared MDP identities within runs second (10,000 replicates). OOD families are analyzed separately rather than pooled. A confidence interval containing zero is reported as unresolved, not as equality or equivalence.

The evidence chronology is also kept separate. The original 1,600-step study met neither predeclared positive nor negative criterion. A subsequent step-count sweep was exploratory, and a 51.2k extension on reused identities was post hoc. The decisive five-run confirmation fixed 51.2k steps, new model/task seeds, and all fixtures before execution; only this fresh block supports the confirmatory repeated-control claim. The similar post-hoc and fresh medians ($1.051$ and $1.052$) are never pooled.

\section{Comparison with learned ICRL/control methods}
The literature-facing benchmark enters only after each external adaptation reproduces a defining behavior of its source method. \textbf{Linear Actor-Critic Transformer} is our implementation of Liang--Lai's one-layer linear-attention actor--critic construction/training. \textbf{Algorithm Distillation} is our causal Transformer trained on source-algorithm histories. AD qualified on a second preregistered qualification run after the first run failed; Liang--Lai's actor step and AD's source actor rate both select the largest value in the predeclared tuning grids. The learned PMD-target reference uses the same architecture, inputs, data, seeds, budget, critic objective, and exact one-step critic as PolicyAttention but is trained directly on Exact PMD targets; it is calibration rather than a headline baseline.

\begin{figure*}[t]\centering\includegraphics[width=.86\textwidth]{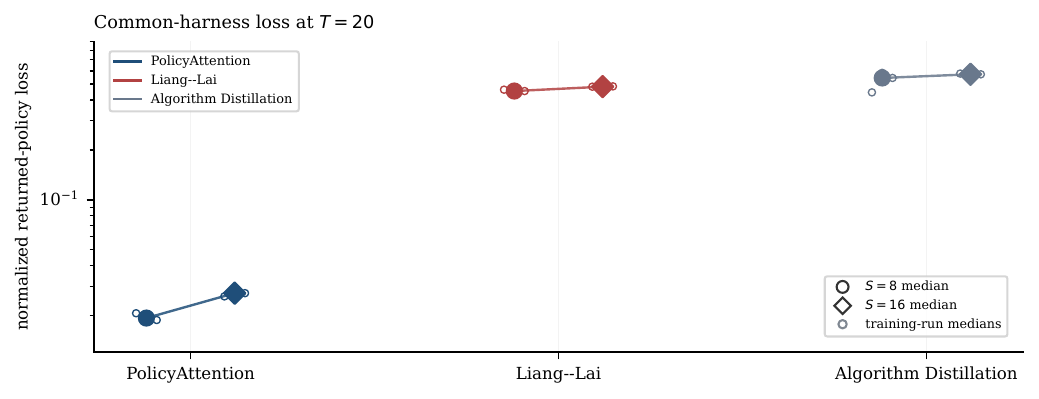}\caption{\textbf{PolicyAttention has substantially lower $T=20$ returned-policy loss than the two qualified external adaptations.} Markers show medians at $S=8,16$ and small open points show trained-run medians. PolicyAttention uses an exact one-step model-based critic here, while the external adaptations receive 20 sampled transitions per round; this is a complete-controller comparison, not an information-matched actor comparison.}\label{fig:lit}\end{figure*}

\begin{table}[t]\centering\scriptsize
\caption{Median normalized returned-policy loss in the common harness (lower is better). Every learned method has three training identities. External entries are our adaptations, not the papers' original benchmarks.}
\begin{tabular}{@{}lrrrr@{}}\toprule
 & \multicolumn{2}{c}{$S=8$} & \multicolumn{2}{c}{$S=16$}\\
Method & $T20$ & $T40$ & $T20$ & $T40$\\\midrule
PolicyAttention & .0193 & .0108 & .0273 & .0224\\
PMD-target ref. (learned) & .0189 & .0106 & .0232 & .0144\\
PolicyAttention, learned critic & .0225 & .0121 & -- & --\\
Liang--Lai adapt. & .4535 & .2812 & .4817 & .3307\\
AD adapt. & .5439 & .3950 & .5723 & .3730\\
\midrule
\emph{Exact PMD oracle} & .0148 & .0026 & .0167 & .0046\\\bottomrule
\end{tabular}
\label{tab:p15}
\end{table}

At $T=20$, PolicyAttention minus Liang--Lai has 95\% intervals $[-0.486,-0.364]$ ($S=8$) and $[-0.496,-0.402]$ ($S=16$); versus AD the intervals are $[-0.612,-0.391]$ and $[-0.620,-0.489]$. Thus PolicyAttention achieves about $18$--$28\times$ lower median loss than the two qualified external adaptations at the tested settings. This is a common-harness separation, not a claim about the original papers and not intrinsic dominance under matched information.

\subsection{Common-harness interfaces}
The learned methods receive different internal information, so the benchmark standardizes task and scoring rather than update information. PolicyAttention and the learned PMD-target reference receive the structured $(\pi,Q,\eta)$ packet and apply one exact model-based Bellman backup after each returned actor policy in the headline controller. Liang--Lai receives a 20-transition on-policy window per round and updates its actor/critic state. AD receives the agent's sampled $(s,a,r)$ history, also 20 new transitions per round, with a sliding 200-token context after qualification. Therefore the $T=20$ comparison gives each external adaptation 400 sampled transitions and gives PolicyAttention model-based one-step evaluation. The resulting gap is a complete-controller comparison and cannot be attributed solely to softmax PMD geometry. At $S=8$, replacing PolicyAttention's exact critic by its learned critic gives median $T=20$ loss $0.0225$ ($1.17\times$ the exact-critic value); Liang--Lai and Algorithm Distillation remain $20.2\times$ and $24.2\times$ higher-loss. This is a one-sided sampled-critic bound, not a matched-information test: the learned critic uses 144 generative transitions per round, versus 20 on-policy transitions per round for each external adaptation.

\begin{table*}[t]\centering\small
\caption{Method interface and training context at $S=8$. ``Exact backup'' is one Bellman policy-evaluation step with model access, not full evaluation to $Q^\pi$. The learned-critic PolicyAttention check uses 144 generative transitions per round and is reported separately in text/Table~\ref{tab:p15}.}
\begin{tabularx}{.97\textwidth}{@{}p{.18\textwidth}p{.21\textwidth}p{.17\textwidth}p{.17\textwidth}X@{}}\toprule
Method & actor/training signal & online access per round & parameters / training data & recorded training cost\\\midrule
PolicyAttention & PMD variational actor & exact one-step backup; 0 samples & 272k; 24 MDPs, 2,048 examples, 51.2k steps & 0.19--0.24 GPU-h\\
PMD-target ref. (learned) & direct Exact-PMD labels & exact one-step backup; 0 samples & same architecture/data/budget as PolicyAttention & 0.19--0.24 GPU-h\\
Liang--Lai adaptation & linear-attention actor--critic teacher mimicking & 20 on-policy transitions & 10.9k; 10,000 MDPs, $10^7$ windows & about 17 CPU-min\\
AD adaptation & action NLL on source learning histories & 20 sampled transitions & 1.14M; 2,000 histories $\times$ 1,000 source steps & 0.63--0.75 GPU-h\\\bottomrule
\end{tabularx}\label{tab:interfaces}
\end{table*}

\subsection{Baseline Qualification and Resources}
The external methods entered the benchmark only after method-specific qualification. For Liang--Lai, the implemented linear-attention block reproduces its analytic actor--critic teacher update to about $10^{-5}$ relative error at the harness shape, and native-shape training tracks the teacher curve. Its actor step $\alpha=2.0$ is selected on a disjoint calibration fixture and lies at the upper edge of the predeclared grid. AD's first registered qualification run failed to improve sufficiently. A single registered repair changed context 800$\to$200, batch 32$\to$128, and training 20k$\to$40k steps; the second attempt improved median normalized loss from $0.821$ to $0.395$ over 40 rounds and qualified. Its source actor rate 4.0 is also the largest predeclared grid value. These choices are disclosed because extending either grid could improve the baselines.

Resources are not matched. At $S=8$, PolicyAttention / PMD-target reference have 272,451 parameters and train on 24 MDPs/2,048 examples for 51,200 steps (about 0.19--0.24 GPU-h per run). Liang--Lai has 10,952 parameters but trains over 10,000 MDPs and $10^7$ windows (about 17 CPU-min per run). AD has about 1.14M parameters and distills 2,000 histories of 1,000 source steps (about 0.63--0.75 GPU-h per run). The benchmark therefore measures returned-policy behavior under a common task/scoring harness, not a compute-normalized efficiency frontier.

The learned PMD-target reference is a calibration control and tells a different story. PolicyAttention and PMD-target reference are unresolved at $S=8$ for both horizons and at $S=16,T=20$. At $S=16,T=40$, PMD-target reference is resolved lower: PolicyAttention--PMD-target reference $=+0.0080$, 95\% interval $[+0.0007,+0.0159]$. We therefore make no objective-superiority claim over the learned PMD-target reference.

\subsection{Qualification of External Adaptations}
Because neither external paper publishes this common benchmark, we first require each adaptation to reproduce a defining behavior of its source method. For Liang--Lai, the explicit linear-attention parameter block is unit-tested against the analytic semi-gradient actor--critic update, and the trained block tracks the same teacher on native $S=9,A=4,\gamma=.5$ prompts before it is moved to the $S=8$ harness. The native trained model reaches relative update error $3.2\times10^{-6}$ and its closed-loop improvement tracks the analytic teacher at 3,000 updates. On the harness shape the update error remains around $1.5\times10^{-5}$. The common benchmark then uses the calibration-selected $(\alpha,\beta)=(2.0,0.2)$ at $S=8$ and $(2.0,0.8)$ at $S=16$; the larger actor step is important because the paper-default values adapt much more slowly at the 20-round common horizon.

AD is qualified differently because its defining phenomenon is improvement from learning-history context rather than imitation of a closed-form update. The final long-context model improves median normalized loss from $0.821$ at round 1 to $0.395$ at round 40 on the disjoint qualification fixture, with 64/64 MDPs improving; a short-context control improves substantially less. The first registered training run failed its qualification criterion and remains part of the audit record. The second attempt was fixed before rerunning and is the only AD model family admitted to the benchmark. At $S=16$ the recipe is retrained rather than independently re-qualified from scratch; the evaluation trajectory still shows in-context improvement from $0.923$ at round 1 to $0.373$ at round 40. These qualifications support the phrase ``our faithful adaptations''; they do not convert our benchmark into a reproduction of either publication's reported numbers.

\subsection{PMD-target calibration}
\begin{figure*}[t]\centering\includegraphics[width=.95\textwidth]{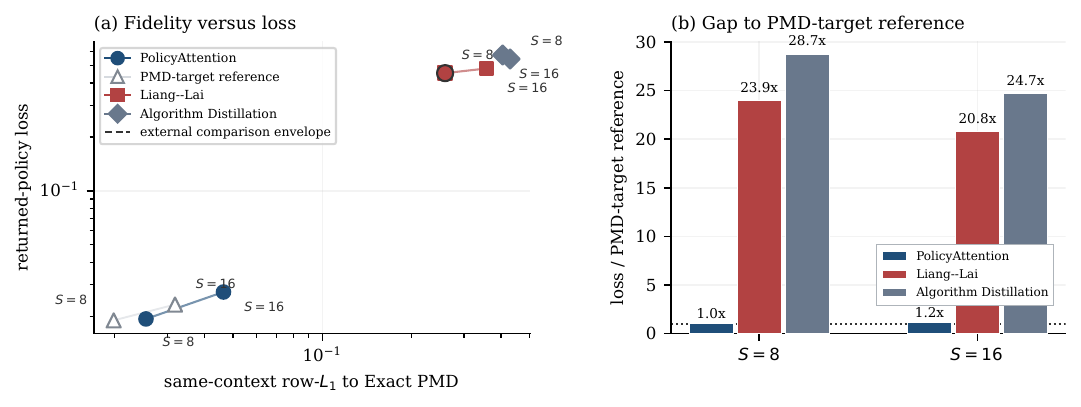}\caption{\textbf{PolicyAttention occupies the low-loss, low-deviation region.} (a) Same-context row-$L_1$ deviation from Exact PMD versus $T=20$ returned-policy loss at $S=8$ and $S=16$; the dashed curve is the non-dominated envelope over the two external adaptations. (b) Common-harness $T=20$ loss normalized by the learned PMD-target reference.}\label{fig:pareto}\end{figure*}

Figure~\ref{fig:pareto}(a) places the $S=8$ and $S=16$ endpoints on a common fidelity--performance plane. PolicyAttention occupies the low-deviation, low-loss region; the Liang--Lai and Algorithm Distillation points form a much higher-loss external comparison envelope. The learned PMD-target reference is shown only as calibration. Figure~\ref{fig:pareto}(b) normalizes $T=20$ loss by that learned reference: PolicyAttention is $1.02\times$ and $1.18\times$ the reference at $S=8$ and $S=16$, while the external ratios inherit the information-access difference of the common harness.

\section{Scope and Boundaries}
Four boundaries constrain interpretation. \textbf{First}, PMD-target reference remains a strong matched control, and one stress-horizon cell favors it. \textbf{Second}, the literature benchmark is not information-matched: PolicyAttention / PMD-target reference use exact one-step model-based backups and zero sampled transitions, while Liang--Lai and AD receive 20 sampled transitions per round; the latter are also plausibly under-tuned because selected rates lie at the edges of the predeclared grids. \textbf{Third}, larger sizes use retraining, not zero-shot state-cardinality extrapolation. \textbf{Fourth}, the theory is an implementation/control result under structured semantic tokens, finite caps, and generative critic coverage; it does not prove SGD/AdamW recovery of the constructed circuit.

Two additional observations sharpen interpretation. A one-state counterexample shows uniformly tiny one-step PMD error can coexist with permanent suboptimal lock-up under bounded steps, motivating the closed-loop test. Separately, a paired reward-only actor achieves lower $S=4,T=20$ task loss than the PMD-aligned actor, and the explicit $\varphi=.01$ support mixture costs $+0.0171$ with 95\% interval $[0.0163,0.0192]$ in $S=4,T=20$ loss in the well-trained regime. These do not negate the PMD mechanism result; they show why algorithmic fidelity and task optimization are different axes.

\section{Discussion}
\paragraph{What the literature-facing advantage establishes.}
The literature-facing comparison is intentionally performance-first: after qualification, every PolicyAttention run has much lower $T=20$ median returned-policy loss than every Liang--Lai and AD run at both tested sizes, and the hierarchical intervals exclude zero by wide margins. The ratio view in Fig.~\ref{fig:lit} makes the magnitude visible without comparing to the oracle: the external adaptations incur $17.7$--$28.2\times$ PolicyAttention's loss. The result survives the increase from eight to sixteen states after retraining. This is the strongest empirical separation in the current project and warrants headline status. Its causal interpretation is narrower. Because the methods expose different information to their update rules, the result establishes that the complete PolicyAttention controller is substantially more effective under this registered common harness; it does not isolate which fraction of the gap comes from actor geometry, exact model-based critic access, optimization, context length, or capacity.

\paragraph{Why the learned PMD-target calibration is complementary.}
The learned PMD-target reference removes most of those differences. It uses the same Transformer architecture, actor inputs, training data, seeds, budgets, and critic objective as PolicyAttention, changing the actor supervision from the proximal variational objective to direct exact-PMD targets. The resulting frontier is mixed rather than one-sided. At $S=8$, the two objectives are unresolved at both horizons and interleave across checkpoints. At $S=16,T=20$, PMD-target reference has a lower point estimate but the interval still crosses zero; at $T=40$ its advantage is resolved. This matched result prevents us from explaining the external-baseline gap as evidence that the proximal objective is intrinsically better than direct imitation. Instead, it supports a mechanism-centered reading: PolicyAttention learns the target computation without direct policy labels and reaches a competitive fidelity/control frontier, while direct supervision can be at least as effective and can win at a longer horizon.

\paragraph{Why local PMD fidelity is still worth measuring.}
Task loss alone cannot identify an update rule. The paired reward-only control demonstrates the point constructively: it attains lower primary-horizon task loss than the PMD-aligned actor while the registered experiment contains no same-context PMD-fidelity coordinate for that reward-only controller. Conversely, the one-state lock-up construction shows that an actor may be arbitrarily close to PMD in one step yet fail under repeated bounded-step control. The combination explains the paper's three-axis evaluation. Same-context $\zeta$ asks which algorithm the actor resembles locally; returned-policy loss asks whether the deployed controller remains useful after feedback; alternative objectives and literature baselines ask how much task performance depends on that mechanism. None of these coordinates is redundant.

\paragraph{What the scale evidence says.}
The $S=8$ and $S=16$ benchmark blocks retrain every method at the new problem size. They therefore test whether the phenomenon persists when the interface is rebuilt at larger tabular state count, not whether one fixed network extrapolates to an unseen cardinality. PolicyAttention's raw median loss grows by $1.41\times$ from $S=8$ to $S=16$ at $T=20$, while the Exact PMD oracle grows by $1.13\times$; its ratio to the oracle therefore changes from $1.31$ to $1.64$. The external adaptations remain far from convergence at the common 20-round horizon, so their raw losses change little across these two sizes. These point ratios are descriptive with three trained runs per method; no uncertainty ordering is inferred for the growth factors.

\section{Conclusion}
PolicyAttention turns the softmax geometry already present in PMD into a causal in-context control mechanism. A fixed decoder realizes the actor--environment--one-step-critic computation with explicit implementation residuals, and the returned-policy theorem connects those residuals to the policy the agent actually emits. Separately trained Transformers exhibit the PMD mechanism locally, preserve useful control under repeated adaptive composition and environment shifts, and, under the common harness, operate in a much lower-loss regime than two qualified published-method adaptations. The decisive distinction is therefore between \emph{algorithmic fidelity}, \emph{closed-loop reliability}, and \emph{task performance}: none can be inferred from the other two. PolicyAttention supplies a common mechanism and measurement interface for studying all three.

\appendix
\section{Complete returned-policy proof}
\label{app:last}
This appendix records the complete error propagation used in the main theorem. Set $E_k=\|Q^\star-Q_k\|_\infty$. Decompose
\begin{align}
Q^\star-Q_{k+1} &= (\F Q^\star-\F Q_k)+(\F Q_k-\F^{\widehat\pi_{k+1}}Q_k)\\
&\qquad -(Q_{k+1}-\F^{\widehat\pi_{k+1}}Q_k).
\end{align}
The first term is bounded by $\gamma E_k$. The second equals $\gamma P g_k$ and has sup norm at most $\gamma\bar g_k$; the third is bounded by $\delta_k$. Thus
\begin{equation}
E_{k+1}\le\gamma E_k+\delta_k+\gamma\bar g_k.
\end{equation}
Unrolling to $T-1$ gives
\begin{equation}
E_{T-1}\le\gamma^{T-1}E_0+\sum_{k=0}^{T-2}\gamma^{T-2-k}\delta_k+\sum_{k=0}^{T-2}\gamma^{T-1-k}\bar g_k.
\end{equation}
For $\pi=\widehat\pi_T$, the Bellman fixed-point equations give
\begin{equation}
Q^\star-Q^\pi=(I-\gamma P^\pi)^{-1}(\F Q^\star-\F^\pi Q^\star).
\end{equation}
The inverse is nonnegative with row sums $1/(1-\gamma)$. At each next state,
\begin{equation}
\max_a Q^\star(s,a)-\ip{\pi(\cdot\mid s)}{Q^\star(s,\cdot)}\le \bar g_{T-1}+2E_{T-1},
\end{equation}
so
\begin{equation}
\|Q^\star-Q^{\widehat\pi_T}\|_\infty\le \frac{2\gamma}{1-\gamma}E_{T-1}+\frac{\gamma}{1-\gamma}\bar g_{T-1}.
\end{equation}
Substitution yields Eq.~(\ref{eq:last}). The proof stops before $Q_T$ is constructed, which is why $\delta_{T-1}$ is absent.

\subsection{From PMD residual to greedy defect}
Let $G_k(s)=\arg\max_aQ_k(s,a)$ and let $p^\circ$ be the current policy restricted to $G_k(s)$ and renormalized. Then $\KL(p^\circ\|\widehat\pi_k)=-\log\widehat\pi_k(G_k(s)\mid s)=D_k(s)$. PMD optimality implies
\begin{equation}
\max_aQ_k(s,a)-\ip{q_k}{Q_k(s,\cdot)}\le D_k(s)/\eta_k.
\end{equation}
For probability rows $p,q$ and $\|Q\|_\infty\le B$, centering $Q$ by the midpoint of its range gives $|\ip{p-q}{Q}|\le B\|p-q\|_1$. Hence
\begin{equation}
g_k(s)\le D_k(s)/\eta_k+B\zeta_k.
\end{equation}
This is the exact interface used by the mixture-floor closure.

\section{Routing and sampling certificates}
\label{app:routing}
The actor partitions the visible memories into the intended state/action group and competitors. If $Z_{\rm good}$ and $Z_{\rm bad}$ are their exponential score sums and $R=Z_{\rm bad}/Z_{\rm good}$, the bad attention mass is $R/(1+R)$. Probability-valued outputs have $L_1$ diameter at most two, giving $\zeta\le2R/(1+R)$. For score margins $\kappa,\nu$, $\|Q\|_\infty\le B$, and actor step $\eta$, the resulting count/score envelope is
\begin{equation}
R_{\rm act}\le(S-1)e^{-\kappa+2\eta B}+Se^{-(\kappa+\nu)+\eta B}.
\end{equation}
For a target $0<\epsilon_r<2$, let $r_\epsilon=\epsilon_r/(2-\epsilon_r)$. A convenient sufficient choice is $\kappa=2\eta_{\max}B+\log(2S/r_\epsilon)$ and $\nu=1$.

The critic has two finite-temperature routing stages. With $M$ visible transition tokens and at least $m$ examples per predecessor,
\begin{align}
R_1&\le(S+A-2)e^{-\tau_1}+(S-1)(A-1)e^{-2\tau_1}+Me^{-3\tau_1},\\
R_2&\le(M/m)e^{-\tau_2}+(2SA/m)e^{-3\tau_2},
\end{align}
and $\delta_{\rm route}\le2B(\gamma\lambda_1+\lambda_2)$ with $\lambda_j=R_j/(1+R_j)$. Conditional on the realized actor and past, fresh TD targets lie in $[-B,B]$ and have conditional mean $\F^{\widehat\pi_{k+1}}Q_k$. Hoeffding plus a union bound over $SAK$ pair/call combinations yields
\begin{equation}
m\ge\left\lceil\frac{2B^2}{\epsilon_s^2}\log\frac{2SAK}{\alpha}\right\rceil
\quad\Longrightarrow\quad \delta_{\rm stat}\le\epsilon_s
\end{equation}
with probability at least $1-\alpha$ over the declared calls. The construction therefore requires packet and step certificates before the routing weights are frozen.

\section{Normalized finite-cap compilation}
\label{app:norm}
Let the normalization-free payload have dimension $d$ and choose $D=d+2$. Let $J\in\mathbb R^{D\times d}$ have orthonormal columns, and choose a unit carrier $c$ orthogonal to the image of $J$ and to the all-ones direction. With $C^2=H^2-D\epsilon_{\rm LN}$ define $y=Cc+Jx$. The token has mean zero and variance $(C^2+\|x\|^2)/D$. LayerNorm with gain $H/\sqrt D$ and zero bias gives
\begin{equation}
J^\top\operatorname{LN}(Cc+Jx)=\frac{Hx}{\sqrt{H^2+\|x\|^2}}=\rho_H(x)x.
\end{equation}
On any finite compact payload domain, $1-\rho_H(x)=O(H^{-2})$. Query/key score perturbations and value perturbations are therefore uniformly small for sufficiently large finite $H$. The final LayerNorm is handled by the same carrier identity.

Action values are represented in centered coordinates $e_a-u$ with $u=\one/A$. The affine readout adds $u$, preserving the simplex. For control we use the explicit mixture
\begin{equation}
\widehat p=(1-\varphi)p+\varphi u,
\end{equation}
not an intrinsic numerical floor from LayerNorm. It guarantees $\widehat p_a\ge\varphi/A$ and adds at most $2\varphi(A-1)/A$ row-$L_1$ actor bias. With constant $\eta=\log(A/\varphi)/\theta$ and uniform residual caps $\delta,\zeta$, the theorem in the main text follows from $\bar g_0\le2B$ and $\bar g_k\le\theta+B\zeta$ for $k\ge1$. The construction is in exact real arithmetic. Conservative sufficient carrier/margin scales can be impractical and are not claimed to match the learned network.

\section{Effective-coordinate learning details}
Let $P_A=I-\one\one^\top/A$ remove the action-constant softmax gauge. In a centered action-equivariant linear class, write effective logits $z_\vartheta=\Phi(x)\vartheta$ and suppose the PMD target is realizable at $\vartheta^\star$. Since the proximal excess equals $\KL(q_\vartheta\|q_{\vartheta^\star})$, differentiation gives
\begin{equation}
\nabla R(\vartheta)=\E[\Phi^\top H(q_\vartheta)\Phi](\vartheta-\vartheta^\star),\quad H(q)=\operatorname{diag}(q)-qq^\top.
\end{equation}
On a finite coefficient ball, bounded features imply bounded logits and hence a positive lower action probability. For centered $v$, $v^\top H(q)v\ge p_0\|v\|^2$. If $\E[\Phi^\top\Phi]\succeq\lambda I$ on the identifiable coordinates, then
\begin{equation}
(\vartheta-\vartheta^\star)^\top\nabla R(\vartheta)\ge p_0\lambda\|\vartheta-\vartheta^\star\|^2.
\end{equation}
Gradient flow contracts on the invariant ball, and sufficiently small-step GD has the analogous geometric convergence. This proposition is intentionally restricted to effective centered coordinates; raw deep-network factorization and AdamW remain outside its scope.

\section{Complete empirical evidence chronology}
\subsection{Fresh confirmation and alternative mechanism}
The decisive fresh block uses five new trained identities at 51.2k steps under a protocol fixed before any outcomes were observed. At $T=20$ the learned/reference run ratios are $0.978413$, $1.396388$, $1.236607$, $0.918064$, and $1.052287$. All five satisfy the $1.5$ registered margin, none systematically stalls, and all certificate/numerical audits pass. The paired reward-only actor produces reward-only-minus-learned-actor median $-0.005963$ with interval $[-0.009794,-0.003191]$, and reward-only-minus-full-loop median $-0.007066$ with interval $[-0.009200,-0.004539]$. These are task-loss comparisons; registered PMD-fidelity diagnostics for the reward-only actor are absent.

The four no-retraining shift medians are $0.9270$ (sparse transitions), $1.0035$ (strong mixing), $1.0138$ (sparse rewards), and $0.9967$ (structured ring). The first three shift families each contain 64 independently generated environments. The ring has one fixed $(P,r)$ structure and 64 initial policies. The aggregate breadth class remains positive if the ring is omitted.

\subsection{Matched objective and checkpoint frontiers}
The matched PolicyAttention / PMD-target reference experiment varies only the actor objective under identical architecture, actor inputs, training fixtures, critic objective, optimizer and checkpoint budgets. At both $S=4$ and $S=8$, the objective comparison is mixed across checkpoints rather than uniformly ordered. The $S=8$ fidelity/control CSV contains five checkpoint budgets from 6.4k to 51.2k. The combined lower-is-better non-dominated envelope contains PolicyAttention at 38.4k $(\zeta=0.01665,L=0.02538)$ and 51.2k $(0.01574,0.02703)$; a PMD-target reference 51.2k point is method-nondominated but dominated in the combined set. These coordinates are descriptive of the registered checkpoint landscape, not evidence that one objective is globally Pareto-superior.

\subsection{Literature-facing benchmark}
At $T=20$, cross-run medians are $0.01929/0.01894/0.45345/0.54390$ for PolicyAttention, PMD-target reference, Liang--Lai, and AD at $S=8$, and $0.02727/0.02321/0.48167/0.57232$ at $S=16$. Figure~\ref{fig:lit} plots the underlying three trained-run medians for every learned method; the trained run remains the independent unit.

At $T=40$, PMD-target reference is resolved lower than PolicyAttention at $S=16$: every PMD-target reference run median (0.0105--0.0158) is below every PolicyAttention run median (0.0208--0.0249). This cell is retained precisely because the external-baseline headline does not determine the matched-control conclusion.

\section{External-baseline qualification details}
The Liang--Lai adaptation uses the paper's one-layer linear self-attention form, actor--critic prompt organization, analytic teacher, and teacher-mimicking training. Native-shape qualification at $S=9,A=4,\gamma=.5$ gives last/first training-loss ratio $4.9\times10^{-12}$, median relative emitted-update error $3.2\times10^{-6}$, closed-loop improvement ratio $0.999$ versus teacher at 3,000 updates, and normalized mean curve gap $4.3\times10^{-4}$. The harness-shape functional error at $S=8$ is $1.5\times10^{-5}$. Calibration selects $\alpha=2.0$, the maximum predeclared grid value, so under-tuning remains a directional caveat.

AD uses causal tokens $(a_{t-1},r_{t-1},s_t)$ and action negative log likelihood, with a side-effect-free returned-policy query. Its first registered qualification attempt ($c=800$, batch 32, 20k steps) failed: round-1/40 losses were 0.928/0.937. After a single registered calibration-only repair to $c=200$, batch 128, 40k steps, the long-context model improves 0.821 $\to$ 0.395 with 64/64 MDPs improved and beats the short-context control by the registered gap. The source actor rate 4.0 is also the maximum predeclared grid value. We therefore use ``qualified adaptation,'' not ``reproduction of Algorithm Distillation's benchmark.''

\section{Negative results and boundaries}
\paragraph{One-step fidelity is insufficient.} For any bounded step cap and target $\zeta>0$, a one-state two-action self-loop can choose sufficiently small good-action prior mass so that exact PMD itself moves by at most $\zeta$. An implemented actor can then stay at the prior while remaining uniformly $\zeta$-accurate to PMD; with an exact critic initialized at $Q^\pi$, the loop remains at a suboptimal fixed point. The aligned proximal excess can simultaneously vanish. This project-specific witness motivates, but does not itself prove, the learned closed-loop result.

\paragraph{Support-floor cost.} In the fresh $S=4$ block, the fixed $\varphi=.01$ mixture has mixture-minus-unmixed-actor median $0.01711$ with interval $[0.01633,0.01919]$: once the actor is well trained, the uniform floor prevents the concentration available to the unmixed actor. The mixture is a theorem-motivated support certificate, not a guaranteed empirical exploration benefit.

\paragraph{Coverage and autonomy.} The construction assumes structured semantic fields and fresh balanced generative critic packets. It does not infer a tabular MDP from raw observations, select state-action coverage endogenously, or support an unlimited growing transcript at fixed finite routing margins. The learned experiments are closed loop in the sense that returned policies change subsequent inputs and targets, but they are not unrestricted autonomous agents.

\paragraph{Size and compute.} $S=16$ uses retraining, not zero-shot cardinality extrapolation. External methods have unmatched capacities and training data; PolicyAttention has stronger online model access. Resource tables are therefore descriptive and not used to claim a compute Pareto optimum.

\section{Reproducibility specification}
The literature benchmark uses three trained identities per method and 64 fresh evaluation MDPs per size. $S=8$ seeds are 18100--18102 for models and 28100--28102 for task roots; $S=16$ uses 19100--19102 and 29100--29102. Calibration/evaluation roots are disjoint. PolicyAttention / PMD-target reference checkpoints use the 51,200-step matched-control runner; external baselines have their own qualified implementations. Exact $Q^\pi$ and $Q^\star$ are used only for scoring (and for the exact one-step critic of PolicyAttention / PMD-target reference). Final intervals use a hierarchical bootstrap with 10,000 replicates, resampling trained runs first and MDP identities within runs second. All manuscript figures are generated deterministically from the compact CSV/JSON inputs shipped with the source package; figure scripts do not train models.

The archived evidence also stores per-round compressed CSV rows, returned-policy arrays, fixture manifests and checkpoint hashes. An independent verification reconstructed the run medians and decisive intervals from those raw rows and rescored stored returned policies with zero discrepancy. The manuscript figures and tables read these fixed evidence artifacts without modifying them.
\bibliographystyle{plainnat}\bibliography{references}\end{document}